\documentclass{IEEE_lsens}
\newcommand{\revision}[1]{#1}
\newcommand{\rev}[1]{#1}

\usepackage{textcomp}
\usepackage[noadjust]{cite}
\usepackage[T1]{fontenc} % optional enhanced font encoding
\usepackage{amsmath}
\usepackage[cmintegrals]{newtxmath}
\usepackage{bm}
\usepackage{array}
\usepackage{url}

\providecommand{\hypersetup}[1]{\relax}

\hypersetup{pdftitle={Bare Demo of IEEE\_lsens.cls for IEEE Sensors Letters},%<!CHANGE!
	pdfsubject={Typesetting},%<!CHANGE!
	pdfauthor={Michael D. Shell},%<!CHANGE!
	pdfkeywords={Class, IEEE, IEEE\_lsens, IEEE Sensors Letters, LaTeX, Typesetting, TeX}}%<^!CHANGE!}

\usepackage[pdftex]{graphicx}
\usepackage{amsmath}
\usepackage{multicol,lipsum}
\usepackage{mathtools}
\usepackage{amsfonts}
\makeatletter
\let\MYcaption\@makecaption
\makeatother

\usepackage[font=footnotesize]{subcaption}

\makeatletter
\let\@makecaption\MYcaption
\makeatother

\usepackage{standalone}

\usepackage[colorlinks,bookmarksopen,bookmarksnumbered,citecolor=red,urlcolor=red]{hyperref} %newcommand{\href}[2]{#2}

\graphicspath{{./figures/}} 

\usepackage{etoolbox}
\apptocmd{\sloppy}{\hbadness 10000\relax}{}{} %To remove underfull hbox warning in bib

\hypersetup %TODO
{
	pdftitle = {On State Estimation for Legged Locomotion over Soft~Terrain}, pdfauthor = {Fahmi, Fink, Semini},
	pdfsubject = {Sensors Letters},
	pdfkeywords = {State Estimation, Legged Locomotion, Soft-Terrain},
	pdftoolbar = true, colorlinks = true,
	linkcolor = black, citecolor = black, urlcolor = black,
}

\usepackage[usenames,dvipsnames]{xcolor}%\usepackage{xcolor,colortbl}
\definecolor{blue_iit}{RGB}{51,51,255}
\definecolor{sfahmi_blue}{RGB}{0.19,0.51,0.74}
\definecolor{LightBlue}{RGB}{0.4,0.4,1}

\usepackage{glossaries}
\newcommand{\sref}[1]{Section~\ref{#1}}
\newcommand{\eref}[1]{(\ref{#1})}
\newcommand{\fref}[1]{Fig.~\ref{#1}}

\newcommand{\ie}{{i.e.},\ }
\newcommand{\eg}{{e.g.},\ }

\newtheorem{assump}{Assumption}

\newcommand{\assref}[1]{Assumption~\ref{#1}}

\makeatletter
\newcounter{definition}

\makeatother

\newacronym{hyq}{HyQ}{hydraulically actuated quadruped}
\newacronym{grfs}{GRFs}{ground reaction forces}

\newacronym{dof}{DoFs}{Degrees of Freedom}

\newacronym{lf}{LF}{left-front}
\newacronym{rf}{RF}{right-front}
\newacronym{lh}{LH}{left-hind}
\newacronym{rh}{RH}{right-hind}

\newacronym{haa}{HAA}{hip abduction/adduction}
\newacronym{hfe}{HFE}{hip flexion/extension}
\newacronym{kfe}{KFE}{knee flexion/extension}

\newacronym{imu}{IMU}{inertial measurement unit}
\newacronym{ekf}{EKF}{extended Kalman filter}
\newacronym{ukf}{EKF}{unscented Kalman filter}
\newacronym{xkf}{XKF}{eXogeneous Kalman Filter}
\newacronym{ltv}{LTV}{linear time-varying}
\newacronym{kf}{KF}{Kalman filter}
\newacronym{ges}{GES}{\revision{globally exponentially stable}}
\newacronym{nlo}{NLO}{non-linear observer}
\newacronym{pe}{PE}{persistency of excitation}
\newacronym{mcs}{MCS}{motion capture system}

\DeclareMathOperator{\Proj}{Proj}
\DeclareMathOperator{\sat}{sat}
\DeclareMathOperator{\vex}{vex}
\DeclareMathOperator{\SO}{SO}

\newcommand{\BF}{\mathcal{B}}
\newcommand{\NF}{\mathcal{N}}

\newcommand{\R}{\mathbb{R}}

\newcommand{\dtau}{\mathop{d\tau}}

\usepackage{tikz}
\usepackage{ifthen}
\usepackage{bm}
\usepackage{rotating}
\usepackage[outline]{contour}
\usepackage[most]{tcolorbox}
\usetikzlibrary{positioning}
\usetikzlibrary{fit}
\usetikzlibrary{calc}
\usetikzlibrary{arrows}
\usetikzlibrary{decorations,decorations.markings,decorations.text}

\newtcbox{\mybox}[1][red]{on line,colback=#1, colframe=#1, boxsep=0pt, boxrule=0pt, size=small, arc=1mm}

\makeatletter
\pgfkeys{/tikz/zdirection/.initial = out}
\pgfkeys{/pgf/decoration/.cd,distance/.initial=1pt}
\newbox\pgfnodepartxlabelbox
\newbox\pgfnodepartylabelbox
\newbox\pgfnodepartzlabelbox
\pgfdeclareshape{axis}
{

\nodeparts{xlabel,ylabel,zlabel}
\anchor{xlabel}{\pgfqpoint{1.0 cm}{-.5\ht\pgfnodepartxlabelbox}}
\anchor{ylabel}{\pgfqpoint{-.5\wd\pgfnodepartylabelbox}{1.0 cm}}
\anchor{zlabel}{\pgfpointadd{\pgfqpoint{-\wd\pgfnodepartzlabelbox}{-\ht\pgfnodepartzlabelbox}}{\pgfqpoint{-0.1cm}{-0.1cm}}}
  \inheritsavedanchors[from=rectangle] 
  \inheritanchorborder[from=rectangle]
  \inheritbackgroundpath[from=rectangle]
  
  \inheritanchor[from=rectangle]{north}
  \inheritanchor[from=rectangle]{north west}
  \inheritanchor[from=rectangle]{north east}
  \inheritanchor[from=rectangle]{center}
  \inheritanchor[from=rectangle]{west}
  \inheritanchor[from=rectangle]{east}
  \inheritanchor[from=rectangle]{mid}
  \inheritanchor[from=rectangle]{mid west}
  \inheritanchor[from=rectangle]{mid east}
  \inheritanchor[from=rectangle]{base}
  \inheritanchor[from=rectangle]{base west}
  \inheritanchor[from=rectangle]{base east}
  \inheritanchor[from=rectangle]{south}
  \inheritanchor[from=rectangle]{south west}
  \inheritanchor[from=rectangle]{south east}
  \inheritanchor[from=rectangle]{base}
  \inheritanchor[from=rectangle]{base west}
  \inheritanchor[from=rectangle]{base east}
  
  \backgroundpath{
	\pgfsetarrowsend{latex}
	\pgfpathmoveto{\pgfpoint{0cm}{0.1cm}}
	\pgfpathlineto{\pgfpoint{0}{1cm}}
	\pgfusepath{stroke}
	\pgfpathmoveto{\pgfpoint{0.1cm}{0cm}}
	\pgfpathlineto{\pgfpoint{1cm}{0cm}}
	\pgfusepath{stroke}
	\pgfsetarrowsend{}
	\ifthenelse{\equal{\pgfkeysvalueof{/tikz/zdirection}}{in}}
           {
          
	  \pgfpathmoveto{\pgfpoint{0.0707cm}{0.0707cm}}
	  \pgfpathlineto{\pgfpoint{-0.0707cm}{-0.0707cm}}
	  \pgfusepath{stroke}
	  \pgfpathmoveto{\pgfpoint{-0.0707cm}{0.0707cm}}
	  \pgfpathlineto{\pgfpoint{0.0707cm}{-0.0707cm}}
	  \pgfusepath{stroke}
	  \pgfpathcircle{\pgfpoint{0cm}{0cm}}{0.1cm}
	  \pgfusepath{stroke}
	  } {
           \pgfpathcircle{\pgfpoint{0cm}{0cm}}{0.1cm}
           \pgfusepath{stroke}
	  \pgfpathcircle{\pgfpoint{0cm}{0cm}}{0.01cm}
	  \pgfusepath{fill,stroke}
	  }

  }
  \foregroundpath{

  }
}

\pgfdeclaredecoration{add dim}{final}{
\state{final}{% 
\pgfmathsetmacro{\dist}{\pgfkeysvalueof{/pgf/decoration/distance}}    
	  \color{blue}
	  \pgfpathmoveto{\pgfpoint{0pt}{0pt}}             
          \pgfpathlineto{\pgfpoint{0pt}{2*\dist}}   
          \pgfpathmoveto{\pgfpoint{\pgfdecoratedpathlength}{0pt}} 
          \pgfpathlineto{\pgfpoint{(\pgfdecoratedpathlength}{2*\dist}}     
          \pgfsetarrowsstart{latex}
          \pgfsetarrowsend{latex}
          \pgfpathmoveto{\pgfpoint{0pt}{\dist}}
          \pgfpathlineto{\pgfpoint{\pgfdecoratedpathlength}{\dist}} 
          \pgfusepath{stroke} 
          \pgfpathmoveto{\pgfpoint{0pt}{0pt}}
          \pgfpathlineto{\pgfpoint{\pgfdecoratedpathlength}{0pt}}
}}
\makeatother

\usepackage{tikz}
\usepackage{ifthen}
\usepackage{etoolbox}
\usetikzlibrary{positioning}
\usetikzlibrary{calc}

\usepackage{pdfpages}

\begin{document}
\markboth{Vol.~XX, No.~XX, XXXX~XXXX}{0000000}
\IEEELSENSarticlesubject{Sensor Applications} % article subject line 

\title{On State Estimation for Legged Locomotion over Soft~Terrain}

\author{\IEEEauthorblockN{
Shamel~Fahmi\IEEEauthorieeemembermark{1}, 
Geoff~Fink\IEEEauthorieeemembermark{2},
and~Claudio~Semini\IEEEauthorieeemembermark{2}}
% <-this % stops a space
\IEEEauthorblockA{Dynamic Legged Systems lab, Istituto Italiano di Tecnologia (IIT), Genova, Italy.\\
\IEEEauthorieeemembermark{1}Student Member, IEEE \hspace{20pt}
\IEEEauthorieeemembermark{2}Member, IEEE}%

% LSENS authors should provide a real e-mail address here.
\thanks{Corresponding author: Shamel Fahmi (e-mail: shamel.fahmi@iit.it).\protect}% <-this % stops a space
\thanks{Associate Editor: XXXX XXXX.}%
\thanks{Digital Object Identifier 10.1109/LSENS.20XX.0000000}}

% Manuscript received line
\IEEELSENSmanuscriptreceived{Manuscript received XXXX; revised XXXX; 
	accepted XXXX. Date of publication XXXX; date of current version XXXX.}

\IEEEtitleabstractindextext{%
\begin{abstract}
Locomotion over soft terrain remains a challenging problem for legged robots. 
Most of the work done on state estimation for legged robots is designed for rigid contacts, and does not take into account the physical parameters of the terrain.
That said, this letter answers the \rev{following} questions: 
how and why does soft terrain affect state estimation for legged robots? 
To do so, 
we utilized a state estimator that fuses \gls*{imu} measurements with leg odometry
that is designed with rigid contact assumptions.
We experimentally validated the state estimator with the HyQ robot trotting
over both soft and rigid terrain. 
We demonstrate that soft terrain negatively affects state estimation for legged robots, 
and that the state estimates have a noticeable drift over soft terrain compared to rigid terrain.
\end{abstract}
\vspace{-5pt}
\begin{IEEEkeywords}
State Estimation, Legged \rev{Robots}, Soft \rev{Contacts}
\end{IEEEkeywords}}

\IEEEpubid{\vspace{0.75cm}\copyright\ 2020 IEEE. Personal use is permitted, but republication/redistribution requires IEEE permission.\\
See \url{http://www.ieee.org/publications\_standards/publications/rights/index.html} for more information.}

%%%%%%%%%%%%%%%%%%%%%%%%%%%%%%%%%%%%%%%%%%%%%%%%%
%%% UNCOMMENT THESE LINES FOR THE COVER PAGE %%%%
\null
\includepdf[pages=-]{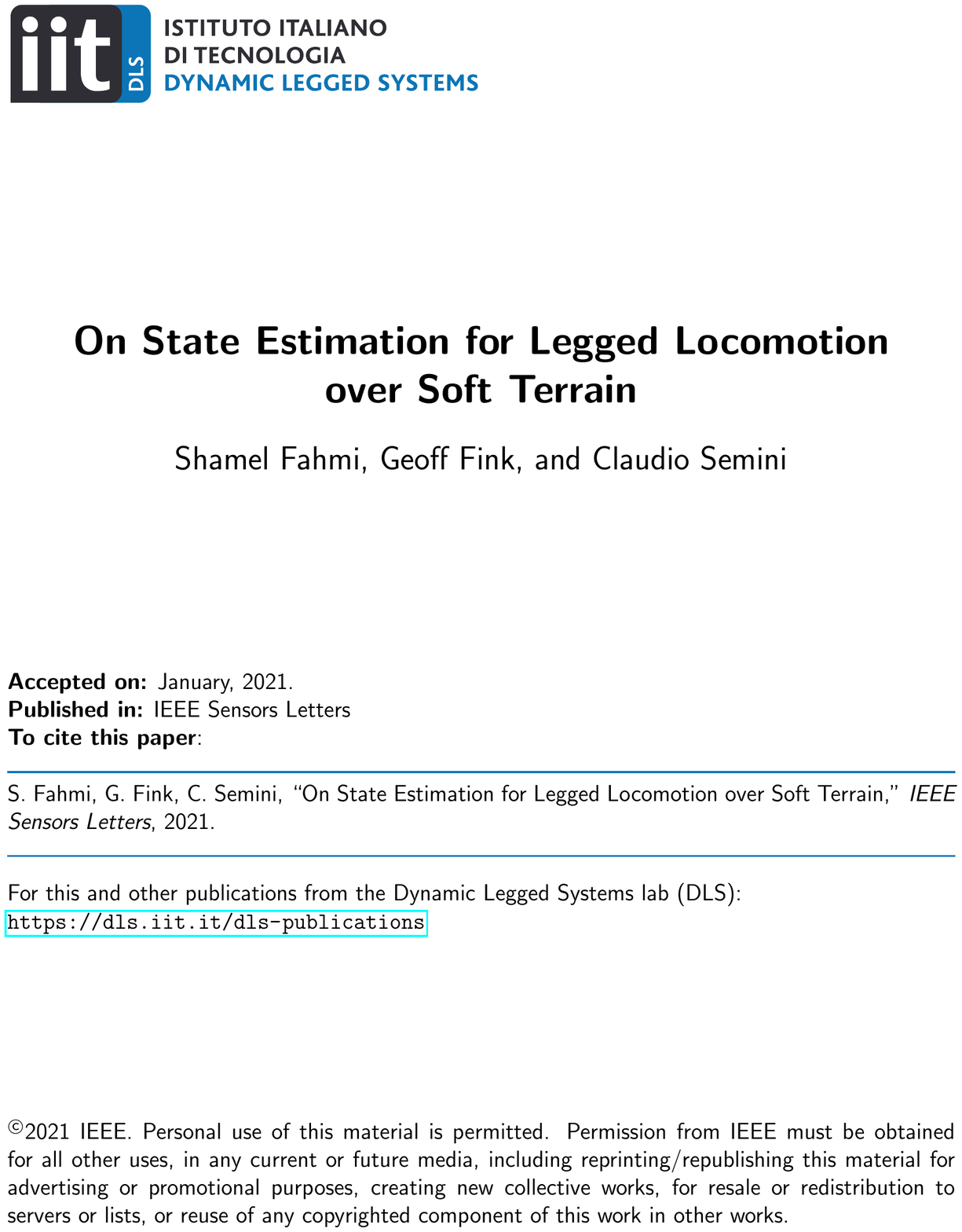}
\setcounter{page}{1}
\maketitle

\section{Introduction}\label{sec_introduction}
Quadruped robots are advancing towards being fully autonomous
as can be seen by their recent development 
in research and industry, and their remarkable agile capabilities
\cite{Semini2019,Bledt2018,Raibert2008}. 
This demands quadruped robots to be robust while 
traversing a wide variety of unexplored complex non-flat terrain.
The terrain may not just vary in geometry, but also in its physical properties 
such as terrain impedance or friction.
Reliable state estimation is a major aspect
for the success of the deployment of quadruped robots
because
most locomotion planners and control strategies
rely on an accurate estimate of the pose and velocity
of the robot. 
Furthermore, reliable state estimation is essential, 
not only for locomotion (low-level state estimation),
but also for autonomous navigation and inspection tasks that are 
emerging applications for quadruped robots (task-level state estimation).

To date, 
most of the work done on state estimation for legged robots 
are based on filters that fuse multiple sensor modalities.
These sensor modalities mainly include 
high frequency inertial measurements and kinematic measurements (\eg leg odometry),
as well as other low frequency modalities (\eg cameras and lidars) to correct the drift.

For instance, 
an \gls*{ekf}-based sensor fusion algorithm has been proposed by~\cite{Nobili2017} 
that fuses \gls*{imu} measurements, leg odometry, stereo vision, and lidar.
In \cite{Ma2016}, a similar algorithm has been proposed that fuses 
\gls*{imu} measurements, leg odometry, stereo vision, and GPS.
In \cite{Bledt2018}, a nonlinear observer 
that fuses \gls*{imu} measurements and leg odometry has been proposed.
In \cite{Fink2020}, a state estimator fuses a \revision{\gls*{ges}} nonlinear attitude
observer based on \gls*{imu} measurements with leg odometry to provide bounded velocity estimates.
The global stability is important for cases when the robot may have fallen over whereas typical 
\gls*{ekf}-based works may diverge. 
The bounded velocity estimates help to decrease drift in the unobservable position estimates. 
Finally, an approach similar to \cite{Fink2020} has been proposed in \cite{Hartley2020}.
This approach proposed an invariant \gls*{ekf}-based sensor fusion algorithm that includes IMU measurements, contact sensor dynamics, and leg odometry. 

\begin{figure}[tb]
	\centering
	\includegraphics[width=0.75\columnwidth]{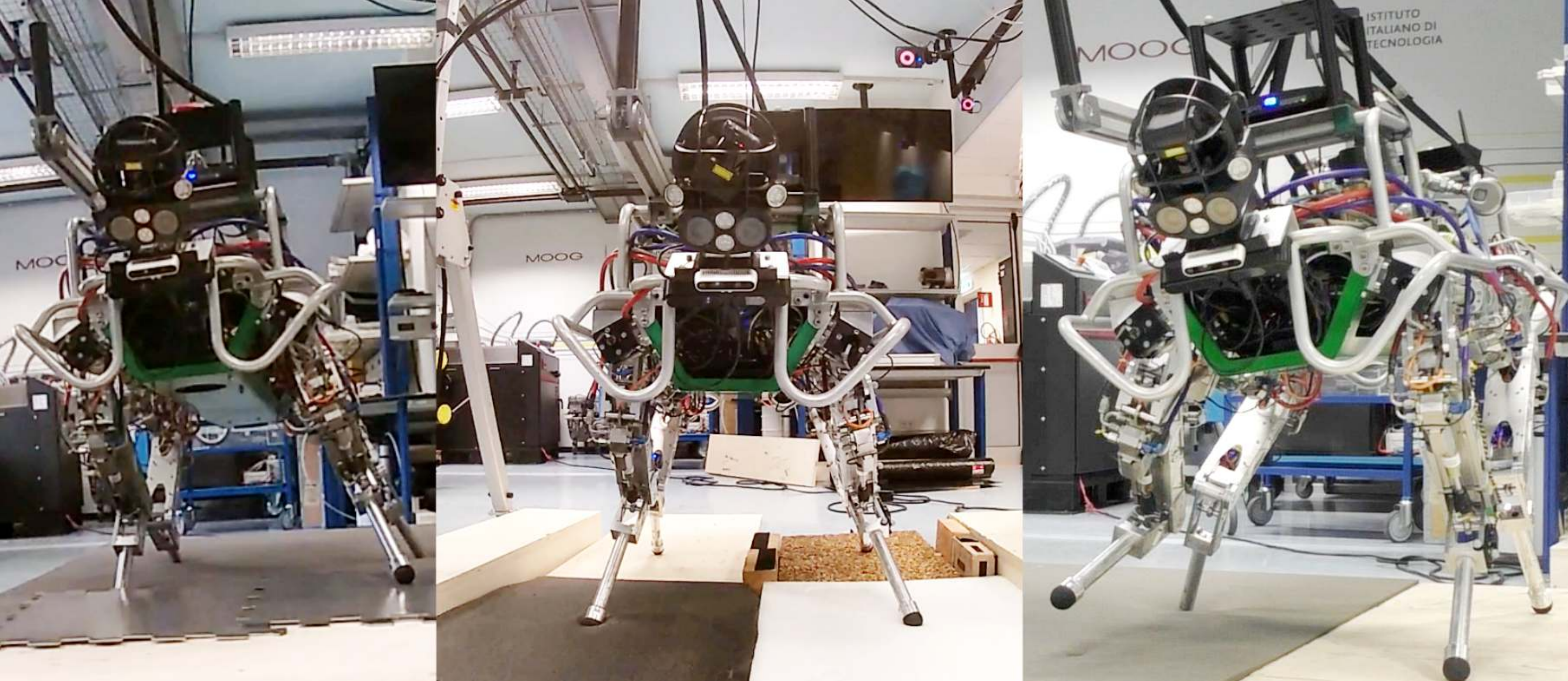}
	\vspace{-5pt}
	\caption{HyQ traversing multiple terrains of different compliances.\label{fig_hyq_soft_terrain}}
	\vspace{-10pt}
\end{figure}

The aforementioned state estimators 
are shown to be reliable on stiff terrain. 
Yet, 
over soft terrain (as shown in~\fref{fig_hyq_soft_terrain}), 
the performance of these state estimators starts to decline. 
\revision{
Over soft terrain,
the state estimator has difficulties 
determining when a foot is in contact with the ground. 
For instance, the state estimator has difficulties determining 
if the foot is in the air, 
if the foot is applying more force than the terrain (terrain compression),
if the terrain itself is applying more force than the foot (terrain expansion), 
or if the foot and the terrain are applying the same force (rigid terrain). 
This results in a large position estimate drift,} and \revision{it} was reported in our previous work~\cite{Fahmi2020} where  
we noted that we encountered difficulties because of state estimation over soft terrain.
Apart from our previous work, 
other works also \revision{mention} that state estimation over soft terrain is a challenging task, 
\eg~\cite{Wish2020, Henze2018}. 
Yet, 
to the authors' knowledge,
literature has not yet discussed the question on how soft terrain affects the state estimation.

\revision{The contributions of this work are
the experimental analysis and formal study on:}
the effects of soft terrain on state estimation, the reasons behind these effects, and simple ways to improve state estimation. 
This letter is building upon our previous work on soft terrain adaptation~\cite{Fahmi2020} 
and on state estimation~\cite{Fink2020}.

The rest of this letter is organized as follows: 
\sref{sec_modeling} describes
the robot model,
the onboard sensors,
and how to estimate the \gls*{grfs} acting on the robot.
\sref{sec_state_est} \revision{explains the state estimator used in this letter, 
and how to estimate the base velocity of the robot using leg odometry}. 
\sref{sec_results} details the results of our experiment and demonstrates how soft terrain affects state estimation.
Finally, \sref{sec_conclusion} presents our conclusions.

\section{Modeling, Sensing, and Estimating}
\label{sec_modeling}

\revision{
In this letter,
we consider the quadruped robot HyQ~\cite{SemTsaGug11} 
shown in~\fref{fig_hyq_soft_terrain}.}
Each leg has three actuated joints.
Despite experimenting on a specific platform, the problem is \revision{generic in nature and it applies equally} to any legged robot.
Furthermore, by using \revision{the $90$~kg HyQ robot}, a~heavy and strong platform, we are exciting more dynamics. %TODO this sentence can be improved

We introduce the following reference frames:
the body frame $\BF$ which is located at the geometric center of the trunk \revision{(robot torso)}, and
the navigation frame $\NF$ which is assumed inertial \revision{(world frame)}.
The basis of the body frame are orientated forward, left, and up.
To simplify notation, the \gls*{imu} is located such that the accelerometer measurements are directly measured in $\BF$.

\noindent\textbf{Kinematics and Dynamics.}
Assuming that all of the external forces are exerted on the feet, the dynamics of the robot is
\begin{equation}\label{eq:fbdynamics}
M(\bar{x})\ddot{\bar{x}}+ h(\bar{x},\dot{\bar{x}})=\bar\tau
\end{equation}
where
$\bar{x}=\begin{bmatrix}x^T&\eta^T&q^T\end{bmatrix}^T\in\R^{18}$ is the generalized robot states,
$\dot{\bar{x}}\in\R^{18}$ is the corresponding generalized velocities,
$\ddot{\bar{x}}\in\R^{18}$ is the corresponding generalized accelerations,
$x\in\R^3$ is the position of the base,
$\eta\in\R^3$ is the attitude of the base,
$q\in\R^{12}$ is the vector of joint angles of the robot,
$M\in\R^{18\times18}$ is the joint-space inertia matrix,
$h$ is the vector of Coriolis, centrifugal and gravity forces,
$\bar\tau=(\begin{bmatrix}0&\tau^T\end{bmatrix}^T-JF)\in\R^{18}$,
$\tau\in\R^{12}$ is the vector of actuated joint torques,
$J\in\R^{18\times 12}$ is the floating base Jacobian, and
$F\in\R^{12}$ is the vector of external forces (\ie \gls*{grfs}).

We solve for the \gls*{grfs} $F_\ell$ of each leg $\ell$ using the actuated part of the dynamics in~\eqref{eq:fbdynamics}.
\begin{equation}\label{eq:grf}
F_\ell = -\alpha_\ell(J_\ell^T(q_\ell))^{-1}(\tau_\ell-h_\ell(\bar{x}_\ell,\dot{\bar{x}}_\ell))
\end{equation}
\mbox{$F_\ell\in\R^3\subset F$} is the \gls*{grfs} for $\ell$ in $\BF$,
\mbox{$J_\ell\in\R^{3\times3}\subset J$} is the foot Jacobian of $\ell$,
\mbox{$\tau_\ell\in\R^3\subset \tau$} is the vector of joint torques of $\ell$,
\mbox{$h_\ell\in\R^3\subset h$} is the vector of centrifugal, Coriolis, gravity torques of $\ell$ in $\BF$,
and 
\mbox{$\alpha_\ell\in\{0,1\}$} selects if the foot is on the ground or not.
A threshold of $F_\ell$ is typically used to calculate $\alpha_\ell$.
\begin{equation}\label{eq:grfalpha}
\alpha_\ell = \begin{cases}
1 & ||(J_\ell^T)^{-1}(\tau_\ell-h_\ell)||>\epsilon \\
0 & \text{otherwise}
\end{cases}
\end{equation}
where $\epsilon>0\in\R$ is the threshold.

\begin{assump}
	There exists a force threshold $\epsilon$ that determines if the foot is in contact with the environment.
	\label{ass1}
\end{assump}

\revision{The translational and rotational kinematics, and the translational dynamics 
of the robot as a single rigid body
in $\NF$ are}
\begin{align}\dot{x}^n=v^n&&\dot{v}^n=a^n+g^n&&\dot{R}^n_b=R^n_bS(\omega^b)\label{eq:pvdyn}\end{align}
where $x^n\in\R^3$, $v^n\in\R^3$, $a^n\in\R^3$ are the position, velocity, and acceleration of the base in $\NF$, respectively, $R_b^n\in \SO(3)$ is the rotation matrix from $\BF$ to $\NF$, and $\omega^b$ is the angular velocity of the base in $\BF$.
The skew symmetric matrix function is $S()$.

\noindent\textbf{Sensors.}
The modeling assumes that the quadruped robot is equipped with 
a six-axis \gls*{imu}  on the trunk
(3~\gls*{dof} gyroscope and 3~\gls*{dof} accelerometer), and that every joint contains an encoder and a torque sensor.
The accelerometer measures specific force $f_s^b\in\R^3$ 
\begin{equation}\label{eq:fs}
f_s^b = a^b + g^b
\end{equation}
where $a^b\in\R^3$ is the acceleration of the body in $\BF$ and $g^b\in\R^3$ is the acceleration due to gravity in $\BF$.
The gyroscope directly measures angular velocity $\omega^b\in\R^3$ in $\BF$.
The encoders are used to measure the joint position $q_i\in\R$ and joint speed $\dot{q}_i\in\R$.
\revision{The pose of each joint (\ie the forward kinematics) is assumed to be exactly known.}
The torque sensors in the joints directly measure torque \revision{$\tau_i\in\R$}.

The measured values of all of the sensors differ from the theoretical values in that they contain a bias and noise: $\tilde{x} = x + b_x + n_x$
where $\tilde{x}$, $b_x$, and $n_x$ are the measured value, bias, and noise of $x$, respectively.
All of the biases are assumed to be constant or slowly time-varying, and all of the noise variables have zero mean and a Gaussian distribution.

\section{State Estimator} \label{sec_state_est}
To compare the effect of different terrains, we use the state-of-the-art low-level state estimator from~\cite{Fink2020}.
It includes input from three proprioceptive sensors: an \gls*{imu}, encoders, and torque sensors.
For reliability and speed no exteroceptive sensors are used.
The state estimator consists of three major components: an attitude observer, leg odometry, and a sensor fusion algorithm.

\noindent\textbf{Non-linear Attitude Observer.}
Typically in the quadruped \revision{robot} literature an EKF is used for attitude estimation, \eg\cite{Nobili2017,Ma2016,Bloesch2013}.
\revision{
However, 
our \textit{attitude observer}~\cite{Fink2020} is \revision{\gls*{ges}}, and it consists of a \textit{\gls*{nlo}}~\cite{Grip2015} and an \textit{\gls*{xkf}}~\cite{Johansen2017}. 
The \gls*{nlo} is
}
\begin{align}\label{eq:NLO}
	\begin{split}
	\dot{\hat{R}}^n_b &= \hat{R}^n_bS(\omega^b-\hat b^b)+\sigma  K_p \rev{J_s}(\hat{R}^n_b)\\ %\Kappa
	\dot{\hat{b}}^b &= \Proj\left(\hat{b}^b,-k\vex\left(\mathbb{P}\left({\hat{R}^{nT}_{bs}} K_p\rev{J_s}(\hat{R}^n_{b})\right)\right)\right)\\ %-\kappa \Kappa
	\rev{J_s}(\hat{R}^n_b) &= \sum_{j=1}^k(y_j^n-\hat{R}^n_by_j^b){y_j^b}^T
	\end{split}
    \end{align}
where $K_p\in\R^{3\times 3}$ is a symmetric positive-definite gain matrix, $k>0\in\R$ is a scalar gain, $\sigma\ge 1\in\R$ is a scaling factor, $\hat{R}^n_{bs}=\sat(\rev{\hat{R}^n_b})$, the function $\sat(X)$ saturates every element of $X$ to $\pm1$,
$\Proj$ is a parameter projection that ensures that $||\hat{b}||<M_b$, $M_b>0\in\R$ is a constant known upper bound on the gyro bias, $\mathbb{P}(X)=\frac{1}{2}(X+X^T)$ for any square matrix $X$, and \rev{$J_s$} is the stabilizing injection term.
The observer is \gls*{ges} for all initial conditions assuming there exists $k>1$ non-collinear vector measurements, \ie $\left| y_i^n\times y_j^n\right|> 0$
where $i,j\in\{1,\cdots,k\}$.
Furthermore, if there is only one measurement the observer is still \gls*{ges} if the following \gls*{pe} condition holds:
\rev{if} there exist constants $T>0\in\R$ and $\gamma>0\in\R$ such that, for all $t\ge 0$,
$\int_t^{t+T}y_1^n(\tau)y_1^n(\tau)^T\dtau\ge \gamma I$
holds then $y_1^n$ is \gls*{pe}.  See~\cite{Grip2015} for proof.

The \gls*{xkf}~\cite{Johansen2017} is similar to an \gls*{ekf} in that it linearizes a nonlinear model about an estimate of the state and then applies the typical \gls*{ltv} \gls*{kf} to the linearized model.
If the estimate is close to the true state then the filter is near-optimal.
However, if the estimate is not close to the true state\rev{,} the filter can quickly diverge.
To overcome this problem\rev{,} the \gls*{xkf} linearizes about a globally stable exogenous signal from a \gls*{nlo}.
The cascaded structure maintains the global stability properties from the \gls*{nlo} and the near-optimal properties from the \gls*{kf}.
The observer is
\begin{align}\label{eq:XKF}
	\begin{split}
	\dot{\hat{x}} &= f_x+\rev{C}(\hat{x}-\breve{x})+K\left(z-h_x-H(\hat{x}-\breve{x})\right)\\
	\dot{P} &=\rev{C}P+P\rev{C^T}-KHP+Q\\
	K&=PH^TR^{-1}
	\end{split}
\end{align}
where $\rev{C}=\left.\partial f_x/\partial{x}\right|_{\breve{x},u}$, $H=\left.\partial h_x/\partial{x}\right|_{\breve{x},u}$, $\breve{x}\in\R^n$ is the bounded estimate of $x$ from the globally stable NLO.
See~\cite{Johansen2017} for the stability proof.

\revision{
\noindent\textbf{Leg Odometry.}
Leg odometry computes the overall base velocity~$\dot x^b$ of the robot by combining the contribution of each foot velocity~$\dot x_{\ell}^b$.
Each leg $\ell$ only contributes to the leg odometry when it is in contact~$\alpha_\ell$.
Thus, we calculate the overall base velocity $\dot x^b$ as}
\begin{equation}
\dot x_{\ell}^b = -\alpha_\ell\left(J_\ell(q_\ell)\dot q-\omega^b \times x_{\ell}^b\right)
\hspace{20pt}
\dot x^b = \dfrac{1}{n_s}\sum_{\ell} \dot x_{\ell}^b
\label{eq:lo}
\end{equation}
where $n_s=\sum\limits_\ell\alpha_\ell$ is the number of stance legs.

\begin{assump}
\revision{	The leg odometry assumes
	that the robot is always in rigid contact with the terrain. 
	This implies 
	\rev{that} the stance feet do not move in $\NF$,
	there is no slippage, 
	the terrain does not expand or compress, 
	and the robot does not jump or fly.}
	\label{ass2}
\end{assump}

\noindent\textbf{Sensor Fusion.}
Lastly, the inertial measurements~\eqref{eq:fs} are fused with the leg odometry~\eqref{eq:lo}. The main advantage of decoupling the attitude from the position and linear velocity is that the resulting dynamics is \gls*{ltv}, and thus has guaranteed stability properties.  \ie the filter will not diverge in finite time.

We use a \gls*{ltv} \gls*{kf} with the dynamics~\eqref{eq:pvdyn}, the accelerometer~\eqref{eq:fs}, and leg odometry~\eqref{eq:lo}.
\begin{align}\label{eq:kf}\begin{split}
\dot{\hat{\underline{x}}}&=f_{\underline{x}}+\underline{K}(\underline{z}-h_{\underline{x}})\\
\dot{\underline{P}} &= \underline{\rev{C}P} + \underline{P\rev{C^T}} -\underline{K}\underline{H}\underline{P}+\underline{Q}\\
\underline{K} &= \underline{P}\underline{H}^T\underline{R}^{-1}
\end{split}\end{align}
where the state $\underline{x}=\begin{bmatrix}{x^n}^T&{v^n}^T\end{bmatrix}^T\in\R^6$ is position and velocity of the base, the input $u=(R_b^nf_s^b-g^n)\in\R^3$ is the acceleration of the base, the measurement $z=R_b^nx_\ell^b\in\R^3$ is the leg odometry, $\underline{K}\in\R^{6\times 3}$ is the Kalman gain, $P\in\R^{6\times 6}$ is the covariance matrix, $Q\in\R^{6\times 6}$ is the process noise and $R\in\R^{3\times 3}$ is the measurement noise covariance, and
\begin{eqnarray*}
f_{\underline{x}}=\begin{bmatrix}v^n\\u\end{bmatrix}\quad
\rev{\underline{C}}=\begin{bmatrix}0_3&I_3\\0_3&0_3\end{bmatrix}\quad
\underline{H}=\begin{bmatrix}0_3&I_3\end{bmatrix}
\end{eqnarray*}
and $I_3$ and $0_3$ are the $3\times 3$ identity matrix and matrix of all zeros, respectively.

\section{Experimental Results} \label{sec_results}
\revision{To analyze the differences in state estimation between rigid and soft terrain, 
we used HyQ and our state estimator.}
HyQ has twelve torque-controlled joints powered by hydraulic actuators.
HyQ has three types of on board proprioceptive sensors:
joint encoders, force/torque sensors, and \revision{\glspl*{imu}}. 
Every joint has 
an absolute and a relative encoder to measure the joint angle and speed. 
The absolute encoder (AMS Programmable Magnetic Rotary Encoder~-~AS5045) 
measures the joint angle when the robot is first turned on, 
while the
relative encoder (Avago Ultra Miniature, High Resolution Incremental Encoder~-~\revision{AEDA-3300-TE1})
measures how far the joint has moved at every epoch. 
Every joint contains a force or torque sensor. Two joints have a load
cell (Burster Subminiature Load Cell~-~8417-6005) and one joint has a custom designed
torque sensor based on strain-gauges. % and is similar to \cite{Khan2017}. 
In the trunk of the robot there is a fibre optic-based, military grade 
KVH 1775 IMU. 

We used the state estimator \eref{eq:NLO}-\eref{eq:kf} on the
\emph{Soft Trot in Place} \revision{and the \emph{Rigid Trot in Place}} dataset from the dataset published in~\cite{Fin19}. 
HyQ was manually controlled to trot on a foam block of $160\times120\times20$~cm, \revision{and on a rigid ground}. 
An indentation test of the foam shows the foam has an average stiffness of 2400~N/m.
All of the sensors were recorded at~1000~Hz.
A \gls*{mcs} recorded the ground truth data with millimetre accuracy at~250~Hz.

\begin{figure}\label{sec_conc}
	\centering
	\newlength{\mylength}
	\def\tikzimagewidth{0.43\columnwidth}\include{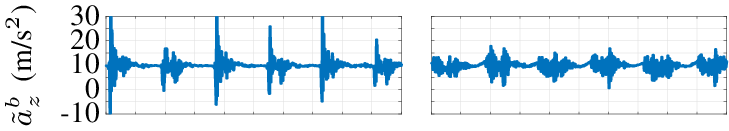}
	\vspace{-0.5cm}\def\tikzimagewidth{0.43\columnwidth}\include{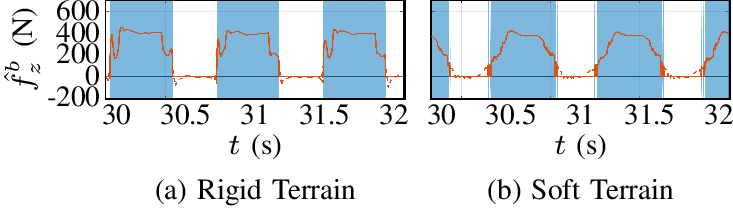}
	\vspace{-0.5cm}\caption{The $z$ component of the measured specific force $\tilde{a}_z^b$ (top),
		and the the estimated ground reaction forces $\hat{f}_z^b$ (bottom),
		in the body frame $\BF$ of HyQ during a trotting experiment.
		The highlighted regions show when the given foot is in stance\rev{,} and the feet are denoted as left-front~(LF), right-front~(RF), left-hind~(LH), and right-hind~(RH).\label{fig:grfz}\label{fig:accz}}
\end{figure}
\def\myfig#1{\begin{minipage}{0.49\textwidth}\centering\def\tikzimagewidth{0.295\columnwidth}\include{figures/#1}\end{minipage}}
\begin{figure*} 
	\centering
	\hspace{-0.04cm}\begin{minipage}{0.49\textwidth}\centering\def\tikzimagewidth{0.295\columnwidth}\include{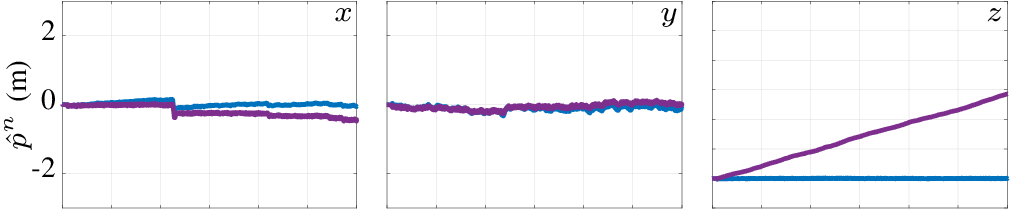}\end{minipage}\quad\hspace{-0.01cm}\begin{minipage}{0.49\textwidth}\centering\def\tikzimagewidth{0.295\columnwidth}\include{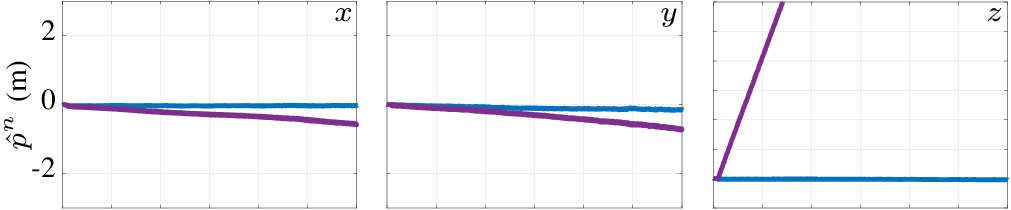}\end{minipage}\\\vspace{-0.3cm}
	\begin{minipage}{0.49\textwidth}\centering\def\tikzimagewidth{0.295\columnwidth}\include{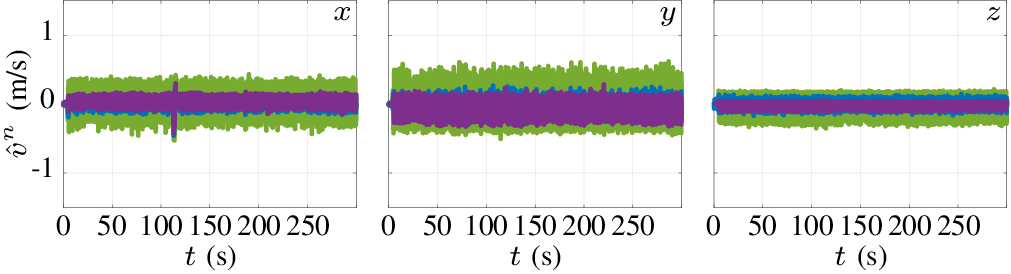}\end{minipage}\quad\begin{minipage}{0.49\textwidth}\centering\def\tikzimagewidth{0.295\columnwidth}\include{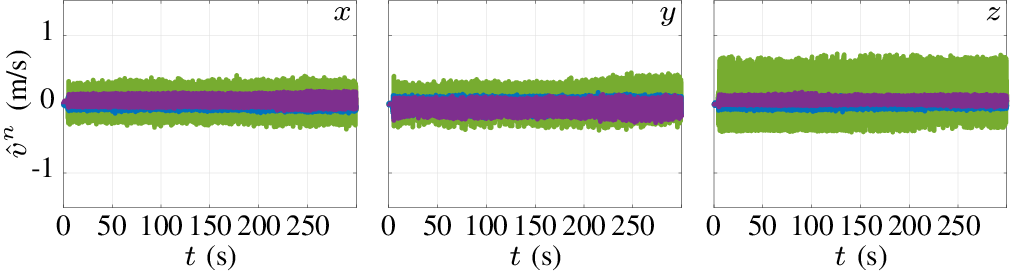}\end{minipage}\\
	\vspace{-0.cm}\begin{minipage}{0.49\textwidth}\centering\def\tikzimagewidth{0.295\columnwidth}\include{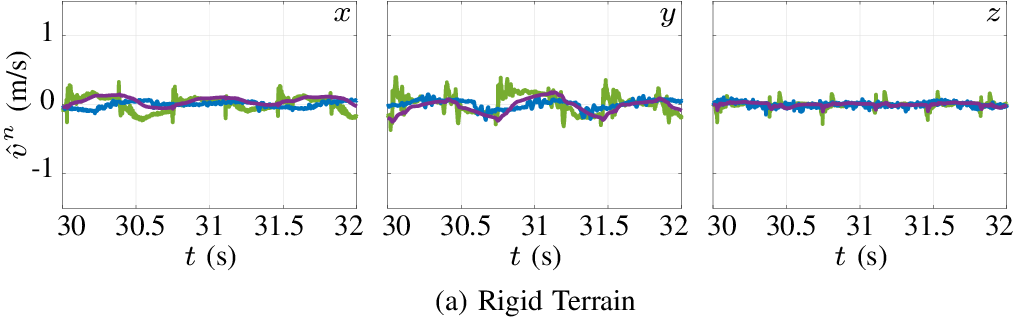}\end{minipage}\quad\begin{minipage}{0.49\textwidth}\centering\def\tikzimagewidth{0.295\columnwidth}\include{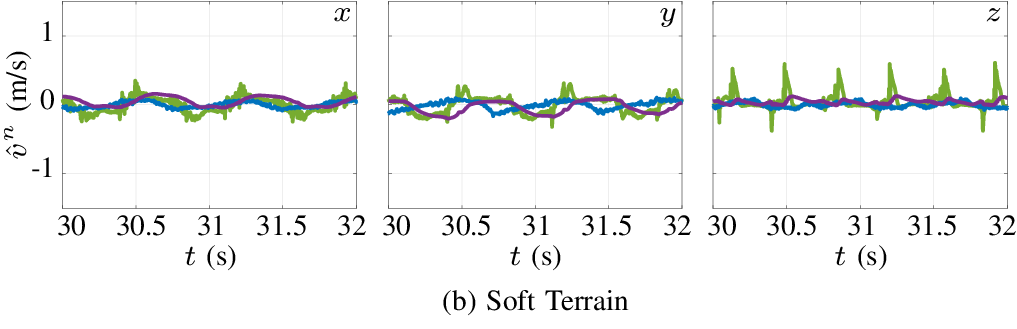}\end{minipage}
	\caption{
		The estimated trunk position $\hat{x}^n$ (top),
		and the estimated trunk velocity $\hat{v}^n$ (middle and bottom),
		in the navigation frame $\NF$ of HyQ during a trotting experiment using sensor fusion (purple) versus the raw leg odometry (green), and the motion capture system (blue).
		The first two rows show the full experiment ($0\le t \le300$)~s and the bottom row is zoomed in ($30\le t \le32$)~s.\label{fig:posvel}}
\end{figure*}

The experiments confirmed our original hypothesis that soft terrain negatively impacts state estimation and also allowed us to investigate why. 
It is important to note that rigid versus soft terrain had no impact on the attitude estimation. 
\revision{For space reasons,} all attitude plots have been omitted.

The first distinct difference between soft and rigid terrain is the specific force measurement of the body as seen in \fref{fig:accz}.  On the rigid terrain there are large impacts and then vibrations every time a foot touches down.  Whereas the soft terrain damped out these vibrations.  Next, on the soft terrain more prolonged periods of positive and negative acceleration can be seen.  This acceleration can also be \revision{seen} in the plots of the \gls*{grfs} in \fref{fig:grfz} where the \gls*{grfs} on the soft terrain are more continuous when compared to the rigid terrain. In other words, there are longer loading and unloading phases. 

The most important differences between soft and rigid terrain are seen in the velocity and position estimates as \revision{shown} in \fref{fig:posvel}. We can see that the leg odometry has large erroneous peaks in $z$ velocity at both touch-down and lift-off.  These peaks in velocities can then be seen in the position estimates as a drift.  
\revision{On the other hand,
the~$x$~and~$y$ position estimates are quite accurate and only have a slow drift.
}

\revision{
In the figures, we can also see multiple of the state estimators assumptions being broken.  
First, there does not exist a constant~$\epsilon$ that can describe when the foot is in contact with the ground, which is contradicting \assref{ass1}.  
The contact~$\epsilon$ is no longer binary 
(\ie supporting/not-supporting the weight of the robot),
but the contact is now a continuous value with varying amounts of the \revision{robot's} weight being supported and sometimes even pushed.  When trying to use the previous simple model, the contact ignores a large portion of the loading and unloading phase.  Furthermore, it often chatters rapidly \rev{between} contact/non-contact when the force is close to~$\epsilon$.  %
Second,
the foot is moving for almost the entire contact (\ie non-zero acceleration) on soft terrain
as shown in \fref{fig:accz}.
This contradicts \assref{ass2} that the foot velocity is zero when in contact.  
Third, \eqref{eq:lo} is broken. It assumes that all of the velocity (and all of the acceleration) is a result of the \gls*{grfs}, but not all of the acceleration due to gravity is being accounted for.  Hence, the robot appears to drift up and away from the ground.
}

There are a few simple ways to try to improve the estimates of this or other similar state estimators. The first is to tune~$\epsilon$ in~\eqref{eq:grfalpha}.  By increasing~$\epsilon$ there would be less erroneous velocity, but in doing so it would also ignore part of the leg odometry.
\revision{In general, 
on a planar surface,
a reduced drift in the~$z$ direction comes at the cost of an increased error in the~$x$ and~$y$ directions.}
A second method could be to have an adaptive velocity bias \revision{for} the leg odometry. However, the bias is not constant and it depends on both the gait and the terrain. Thus, the problem of estimating the body velocity of the robot using leg odometry remains open.

\section{Conclusions}\label{sec_conclusion}
In this letter,
we present an experimental validation and a formal study on  
the influence of soft terrain on state estimation for legged robots.
We \revision{utilized} a \revision{state-of-the-art} state estimator that fuses \gls*{imu} measurements with leg odometry. 
\revision{We experimentally analyzed the differences between soft and rigid terrain using our state estimator and a dataset of the HyQ robot.}
\rev{That said, we report three main outcomes.}
\rev{First, w}e showed that soft terrain results in a larger drift in the position estimates,
and larger errors in the velocity estimates compared to rigid terrain. 
These problems are caused by the broken legged odometry contact assumptions on soft terrain.
\rev{Second, w}e also showed that over soft terrain, the contact with the terrain is no longer binary
and it often chatters rapidly between contact and non-contact.
\rev{Third, we showed that soft terrain 
	affects many states besides the robot pose. 
This includes the contact state and the \gls*{grfs} which are essential for the control of legged robots.}
Future works include extending the state estimator to incorporate the terrain impedance in the leg odometry model. 
Additionally, further datasets will be recorded to investigate the \revision{long-term} drift in 
the forward and lateral directions.

\vspace{-10pt}
%\bibliographystyle{./includes/IEEEtran}
%\bibliography{includes/references.bib}

%

\end{document}